\documentclass[10pt]{ieeetran} 
\usepackage[left=54pt, right=54pt, bottom=54pt, top=72pt]{geometry}
\title{An Integrated Real-time UAV Trajectory Optimization with Potential Field Approach for Dynamic Collision Avoidance}
\author{D. M. K. K. Venkateswara Rao, Hamed Habibi \textit{Member, IEEE}, Jose Luis Sanchez-Lopez, and Holger Voos
\thanks{This research was partially supported by the European Union’s Horizon 2020 project Secure and Safe Multi-Robot Systems (SESAME) under the grant agreement no. 101017258 and by the Department of Media, Telecommunications and Digital Policy (SMC) of the Government of the Gran Duchy of Luxembourg under the project reference SMC/CFP-2019/010/IRANATA, "Interference and RAdiation in Network PlAnning of 5G AcTive Antenna Systems."}
 \thanks{D. M. K. K. Venkateswara Rao, H. Habibi, J. L. Sanchez-Lopez, and H. Voos are with Automation and Robotics Research Group, Interdisciplinary Centre for Security, Reliability and Trust, University of Luxembourg, Luxembourg. H. Voos is also with Faculty of Science, Technology and Medicine (FSTM), Department of Engineering, University of Luxembourg, 
{\tt\small \{mohan.dasari,hamed.habibi,holger.voos,\\joseluis.sanchezlopez\}@uni.lu}
}
}

\date{}

\usepackage{amsthm}
\usepackage{amssymb}
\usepackage{amsmath}
\usepackage{breqn}
\usepackage{hyperref}
\hypersetup{
    colorlinks=true,
    linkcolor=blue,
    citecolor=red,
}
\usepackage{color,soul}
\usepackage{graphics} 
\usepackage{epsfig}
\usepackage{mathptmx} 
\usepackage{times}
\usepackage{amssymb}
\usepackage[mathscr]{euscript}
\usepackage{enumerate}
\usepackage{algpseudocode}
\usepackage{algorithm}
\usepackage{physics}
\usepackage{multirow}
\usepackage{optidef}

\DeclareMathAlphabet{\mbf}{OT1}{ptm}{b}{n}

\newcommand{\argmin}{\operatornamewithlimits{argmin}}
\graphicspath{{figures/}}
\begin{document}
\maketitle
\begin{abstract}
This paper presents an integrated approach that combines trajectory optimization and Artificial Potential Field (APF) method for real-time optimal Unmanned Aerial Vehicle (UAV) trajectory planning and dynamic collision avoidance. A minimum-time trajectory optimization problem is formulated with initial and final positions as boundary conditions and collision avoidance as constraints. It is transcribed into a nonlinear programming problem using Chebyshev pseudospectral method. The state and control histories are approximated by using Lagrange polynomials and the collocation points are used to satisfy constraints. A novel sigmoid-type collision avoidance constraint is proposed to overcome the drawbacks of Lagrange polynomial approximation in pseudospectral methods that only guarantees inequality constraint satisfaction only at nodal points. Automatic differentiation of cost function and constraints is used to quickly determine their gradient and Jacobian, respectively. An APF method is used to update the optimal control inputs for guaranteeing collision avoidance. The trajectory optimization and APF method are implemented in a closed-loop fashion continuously, but in parallel at moderate and high frequencies, respectively. The initial guess for the optimization is provided based on the previous solution. The proposed approach is tested and validated through indoor experiments.
\vspace*{3mm}

Experiment video link: \url{https://youtu.be/swSspfvYjJs}
\end{abstract}


\section{Introduction}\label{intro}
Trajectory planning is one of the most important features for autonomous Unmanned Aerial Vehicle (UAV) in cluttered environments, defined as a time-parameterized motion reference, i.e., geometric values of position, heading, derivatives associated with time law, passing through the waypoints, considering geometrical feasibility, collision avoidance, kinematics and dynamics \cite{castillo2020real}. One of the main objectives of the trajectory planner is collision avoidance with obstacles, humans, and other robots. These, generally, can be titled as static and dynamic obstacles \cite{ sanchez2020trajectory}. So, the information on the obstacles is to be implemented into the proposed solution. Collision avoidance is considered as the geometrical feasibility of the planning. 
Most of the planning approaches are constructed as an on/offline optimization problem, obtaining a trajectory with the optimized feature \cite{gao2019flying}. Mostly, the time that the UAV takes between two given points is minimized, aiming at the agile maneuver. More importantly, the real-time implementation is more desirable for navigation in dynamic environments \cite{zhou2020robust, zhou2021ego}. This implies the need for an algorithm that converges to a feasible solution with a low computational burden. In terms of technical development, the representation of the optimization problem with the considered constraints usually determines the reliability of the solution in practice. 

Trajectory optimization has been widely investigated in the last two decades \cite{hoy2015algorithms}. Hard-constrained methods are pioneered, in which piecewise polynomial trajectories are generated through quadratic programming \cite{richter2016polynomial}. Free space can be represented by a sequence of cubes \cite{chen2015real}, spheres \cite{gao2019flying} or polyhedrons \cite{liu2017planning}. Improper time allocation of polynomials, e.g., naive heuristics, leads to unsatisfying results. Fast marching \cite{gao2018online} and kinodynamic \cite{ding2019efficient, gao2018online} are used to find a feasible initial guess, ensuring global optimality by the convex formulation. However, distance to obstacles in the free space is ignored, which often results in trajectories being close to obstacles. Moreover, kinodynamic constraints are conservative, making the trajectory’s speed deficient for fast flight \cite{tordesillas2021faster}. A feasible solution can only be obtained by iteratively adding more constraints and solving the quadratic programming problem, which is undesirable for real-time applications.

Soft-constrained methods translate the problem to a non-linear optimization problem with smoothness and safety \cite{zucker2013chomp}. Since the time parameterization is continuous, it avoids numeric differentiation errors, with more accuracy to represent motions. However, it suffers from a low success rate. As a resolution, in \cite{gao2017gradient} a feasible initial path is found using an informed sampling-based path-searching method. In \cite{usenko2017real}, uniform B-spline is used for the trajectory parameterization, as B-spline is continuous by nature, the continuity constraint is avoided. Moreover, it is beneficial for local re-planning, due to its locality property. Soft-constrained methods might stick in local minima and cannot guarantee of success rate and kinodynamic feasibility. 

Gradient-based method is the mainstream approach \cite{ zhou2020robust, zhou2021ego}, as a non-linear optimization, taking into account smoothness, feasibility, and safety using various parameterization methods, including polynomial and B-spline. The computation time can be reduced by compact environment representation \cite{zhou2020ego}. Furthermore, it is effective for local re-planning, which is useful for high-speed flight in unknown environments. However, the local minima are still an issue. The stochastic sampling strategy partially overcomes this issue with the cost of computationally intensive \cite{oleynikova2016continuous}. The quality of the initial guess can improve the success rate \cite{gao2017gradient}. \cite{ gao2020teach} used an iterative post-process to improve the success rate of \cite{ zhou2019robust}. 

Topological path planning method utilises the idea of topologically distinct paths for planning, in which paths belonging to different homotopy are used to escape local minima. Trajectory planning is presented in \cite{ rosmann2017integrated} in distinctive topologies using Voronoi and sampling-based front-ends and Timed-Elastic-Bands local planner \cite{ rosmann2012trajectory} as back-ends. Based on \cite{ jaillet2008path}, in \cite{zhou2020robust} a real-time topological planning is proposed by efficient topology equivalence checking. In most of the similar works, collision avoidance in dynamic environments is considered, considering the velocity of obstacles \cite{ alonso2015collision}, decentralized Nonlinear Model Predictive Control (NMPC) \cite{ shim2003decentralized} and sequential NMPC \cite{ nageli2017real}. Uncertainties can be handled by enlarging bounding volumes \cite{ gopalakrishnan2017chance}, which might lead to conservative or infeasible solutions. Moreover, the chance-constrained approaches are computationally intensive and thus not eligible for real-time collision avoidance \cite{blackmore2010probabilistic}.

Motivated by the above-mentioned considerations, in this paper we design an integrated solution for trajectory optimization with the Artificial Potential Field (APF), satisfying real-time minimum-time UAV trajectory planning and dynamic collision avoidance. 
The problem is transcribed into a nonlinear programming problem using Chebyshev pseudospectral methods. The state and control histories are approximated by using Lagrange polynomials and collocation is used to satisfy the dynamics constraints. The obstacle avoidance constraint is modelled by a novel sigmoid function to overcome the drawbacks of Lagrange polynomial approximation in pseudospectral methods that only guarantees inequality constraint satisfaction only at nodal points. It also guarantees the feasibility of the optimization problem. Furthermore, to reduce the computational complexity, automatic differentiation is used to obtain the gradient of cost function and Jacobian of constraints. An APF method is used to update the optimal control inputs for guaranteeing collision avoidance. The proposed framework structure can overcome convergence issues, by having the components run in parallel but at moderate and high frequencies.

The rest of the paper is organized as follows. The UAV dynamics model, optimization formulation, and NLP transcription are covered in Sec.~\ref{sec:formulation}. Sec.~\ref{sec:collision_avoidance} comprises the novel constraint formulation for collision avoidance, the APF method for guaranteed collision avoidance, and computational implementation. The proposed architecture is also presented.  Indoor lab setup, and experimental results are presented in \ref{sec:results}. Finally, the paper ends with conclusions in Sec. \ref{sec:conclusions}.

\section{Problem Formulation and Preliminaries} 
\label{sec:formulation}
\subsection{Dynamics Model}
Consider an approximate dynamics model for the UAV as a first-order dynamics system in terms of the position vector $\underline{x}(t)=[x(t),y(t),z(t)]^T \in \mathbb{R}^3$ and commanded position vector $\underline{u}_{cmd}(t)=[x_{cmd}(t),y_{cmd}(t),z_{cmd}(t)]^T \in \mathbb{R}^3$, both presented in the inertial frame of reference, as
\begin{equation} \label{eqn:uav_dynamics}
\dot{\underline{x}}(t)=diag(K_x,K_y,K_z)\left(\underline{u}_{cmd}(t)-\underline{x}(t)\right),
\end{equation}
where $K_x$, $K_y$, and $K_z$ are constant dynamics gains.

The model \eqref{eqn:uav_dynamics} assumes that the position of the UAV converges to the commanded position asymptotically \cite{castillo2020real}. The low-level controller in the autopilot computes the required motor inputs based on the error between current and commanded positions, with desirable performance. 

\subsection{Optimization Problem Formulation}
The trajectory planning problem is to determine the control inputs required to transfer the UAV from the given initial to final positions in minimum time. 
The trajectory planning problem is formulated to determine optimal position vector $\underline{x}_{opt}(t)=[x_{opt}(t),y_{opt}(t),z_{opt}(t)]^T \in \mathbb{R}^3$ and corresponding commanded position vector $\underline{u}_{cmd}(t)$, such that to steer $\underline{x}(t)$ towards $\underline{x}_{opt}(t)$, in minimum time, starting from the initial position $\underline{x}(t_0)=\underline{x}_0 \in \mathbb{R}^3$ to the final position $\underline{x}(t_f)=\underline{x}_f \in \mathbb{R}^3$, where $t_f$ is the unknown terminal time. Also, it is aimed to avoid $n_o$ moving obstacles with the position vector $\underline{x}_{obs,k}(t)=[x_{obs,k}(t),y_{obs,k}(t),z_{obs,k}(t)]^T \in \mathbb{R}^3$, for $k \in \{1,...,n_o\}$, by keeping the UAV outside of safety spheres, considered around the obstacles. Furthermore, there are $n_w$ waypoints that UAV is supposed to pass through before reaching $\underline{x}_f$. 
This optimization problem is mathematically formulated as
\begin{equation}
    [\underline{x}_{opt}(t),\underline{u}_{cmd}(t)]= \argmin_{\underline{u}_{cmd}(t)} {J\left(x(t),\underline{u}_{cmd}(t),\underline{x}_0,\underline{x}_f\right)}, 
    \label{eqn:cost_function}
\end{equation}
where
\begin{equation}
    J(\cdot)= t_f\left(\underline{x}(t),\underline{u}_{cmd}(t),\underline{x}_f\right)-t_0,
\end{equation}
subject to dynamics \eqref{eqn:uav_dynamics} and
boundary conditions
\begin{subequations}
\begin{align}
    \underline{x}(t_0)&=\underline{x}_0,\\
    \underline{x}(t_f)&=\underline{x}_f,\\
    \underline{x}(t_j)&=w_{j},
\end{align}    
\label{eqn:boundary_conditions}
\end{subequations}
bound constraints on position and speed 
\begin{subequations}
    \begin{align}
        \underline{x}_L \leq \underline{x}(t) \leq \underline{x}_U\\
        \underline{u}_L \leq \underline{u}_{cmd}(t) \leq \underline{u}_U\\
        \underline{\dot{x}}_L \leq \underline{\dot{x}}(t) \leq \underline{\dot{x}}_U
    \end{align}
    \label{eqn:bound_constraints}
\end{subequations}
and obstacle avoidance constraints
\begin{align}
    R_k \leq \norm{\underline{x}(t)-\underline{x}_{obs,k}(t)},   \label{eqn:obstacle_constraints}
\end{align}
for $j \in \{1,...,n_w\}$ and $k \in \{1,...,n_o\}$, where, $t_0\leq t_j \leq t_f$, are increasing time sequence, and $w_j=[x_{j}(t),y_{j}(t),z_{j}(t)]^T \in \mathbb{R}^3$ is the position of the $j^{th}$ waypoint. $X_L$ and $X_U$ denote the element-wise lower and upper bound vectors on the variable vector $X(t)$. Finally, $R_k$ for $k \in \{1,...,n_o\}$, represents safety radius of $k^{th}$ obstacle. In fact, $ R_k \leq \norm{x(t)-x_{obs,k}(t)}$ imposes the constraint to keep the position of the UAV outside of the safety sphere around the obstacle.
\subsection{NLP Transcription:}
\begin{figure}[thpb]
\centering
\includegraphics[width=90mm]{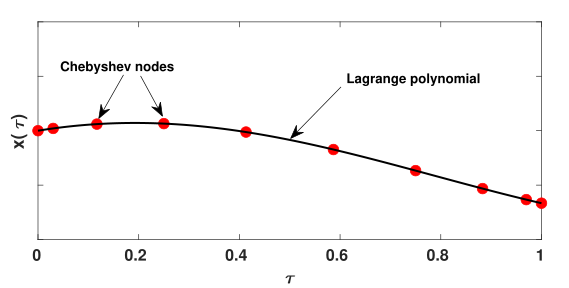}
\caption{Illustration of trajectory approximation using Chebyshev pseudospectral method.}
\label{fig:chebyshev_psm}
\end{figure}

The continuous-time trajectory optimization problem \eqref{eqn:cost_function} is transcribed into a nonlinear programming (NLP) problem using Chebyshev pseudospectral method~\cite{fahroo2002direct}. In this method, the state and control histories between the initial and final waypoints are approximated by Lagrange polynomials in non-dimensionalized time, as illustrated in Figure~\ref{fig:chebyshev_psm}

\begin{subequations}
\begin{align}
{\underline{x}}(\tau) \approx \sum_{i=1}^{n} {\underline{x}}_{i}\phi_i(\tau),\\
{\underline{u}}_{cmd}(\tau) \approx \sum_{i=1}^{n} \underline{u}_{i}\phi_i(\tau),
\end{align}
\end{subequations}
where $\tau \in [0,1] $ is the non-dimensionalized time; $n$ is the order of the polynomial; $\underline{x}_i$ and $\underline{u}_i$ are state and control approximation vectors at nodes; $i$ denotes node number; and $\phi_i(\tau)$ is the basis function given by
\begin{align}
\phi_i(\tau) = \prod_{j=1,j\neq i}^{n}\frac{\tau-\tau_j}{\tau_i-\tau_j},
\end{align}
where $\tau_i$ and $\tau_j$ are roots of shifted Chebyshev polynomials obtained by
\begin{align}
\tau_i = \frac{1+\tilde{\tau}_i}{2}.
\end{align}
The roots of the Chebyshev polynomial are determined as
\begin{align}
    \tilde{\tau}_i = \cos\left(\pi k/n \right),~~~ k= 0,1,...,n 
\end{align}
Let $\Delta t$ be the maneuver time. The actual and non-dimensionalized times are related as
\begin{align}
\tau & = \frac{t}{\Delta t},\\
d\tau & = \frac{dt}{\Delta t}.
\label{eqn:t_tau_relation}
\end{align}
The Lagrange polynomials are collocated at the interpolation points to result in a set of algebraic equations
\begin{equation}
\underline{x}_{i}\dot{\phi}_i(\tau) = \dot{\underline{x}}(\tau).
\end{equation}
This can be rewritten as
\begin{equation}
\sum_{i=1}^{n}\underline{x}_{i}\dot{\phi}_i(\tau) = \Delta t f(\underline{x}_i,\underline{u}_i).
\label{eqn:nlp_col_constraints}
\end{equation}
Then, the cost function $J$, in \eqref{eqn:cost_function} reduces to 
\begin{align}
J = \Delta t.
\label{eqn:J_psm}
\end{align}

The problem \eqref{eqn:cost_function} with cost function \eqref{eqn:J_psm} with boundary conditions \eqref{eqn:boundary_conditions}, bound constraints \eqref{eqn:bound_constraints}, and obstacle avoidance constraints \eqref{eqn:obstacle_constraints}, constitute the NLP problem. This NLP problem is solved using interior point algorithm \cite{byrd1999interior} implemented in IPOPT solver \cite{wachter2006implementation} interfaced with Matlab. Moreover, CasADi~\cite{andersson2019casadi} is used for the automatic differentiation to determine the gradient and Jacobian of cost function and constraints, respectively, for speeding up the computation of the solution, which is vital for real-time implementation.

\section{Collision Avoidance}
\label{sec:collision_avoidance}
\begin{figure}[thpb]
\centering
\includegraphics[width=85mm]{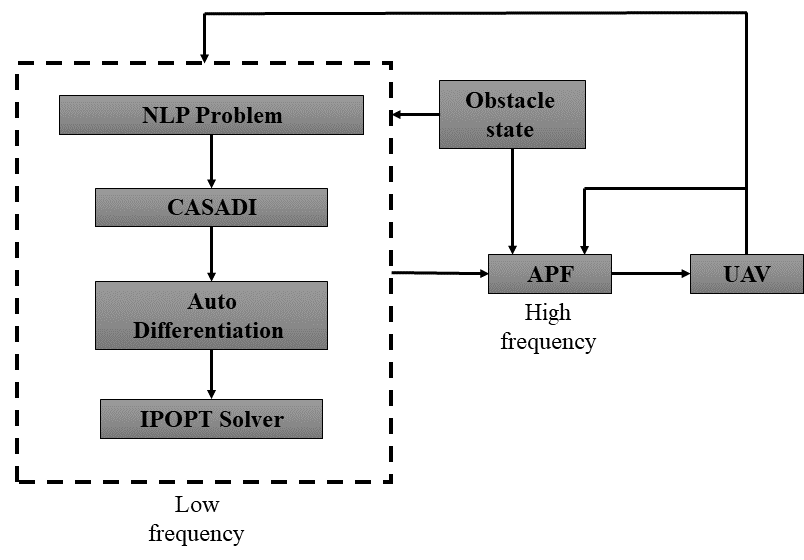}
\caption{Computational architecture.}
\label{fig:comp_arch}
\end{figure}

In this section the modifications applied to the optimization problem \eqref{eqn:J_psm}, are addressed that guarantee the feasibility of the problem and the fast convergence of the solver, aiming at the real-time re-optimization. The overall computational architecture is presented in Figure \ref{fig:comp_arch}.

\subsection{Constraints Formulation}
\begin{figure}[!htbp]
\centering
\includegraphics[scale=0.7]{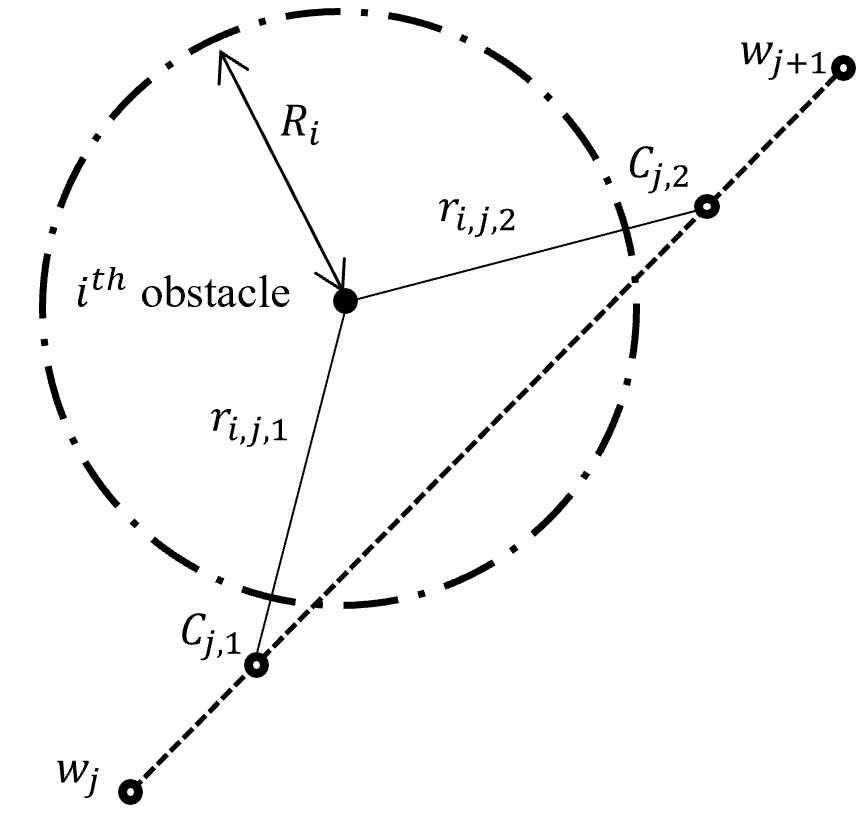}
\caption{Violation of safety region for the case the waypoints and collocation points are still outside.}
\label{fig2}
\end{figure}

To apply the obstacle avoidance constraint \eqref{eqn:obstacle_constraints} to the collocation points, the common condition to be checked is the distance of the points from $i^{th}$ obstacle to be greater than the safety radius $R_i$ for $i \in \{1,...,n_o\}$. However, this might lead to the safety violation as illustrated in Figure \ref{fig2}. Evidently, the collocation points $C_{j,1}$ and $C_{j,2}$ are outside of the safety sphere, i.e., $r_{i,j,1}>R_i$ and $r_{i,j,2}>R_i$. However, the line segment $\overline{C_{j,1}C_{j,2}}$ is still passing through the safety sphere. The trivial solution is to increase the number of collocation points and apply the distance constraint. However, this increases the computational burden. More importantly, it is still not geometrically guaranteed that connecting lines are moved outside. 

\begin{figure}[!htbp]
\centering
\includegraphics[scale=0.7]{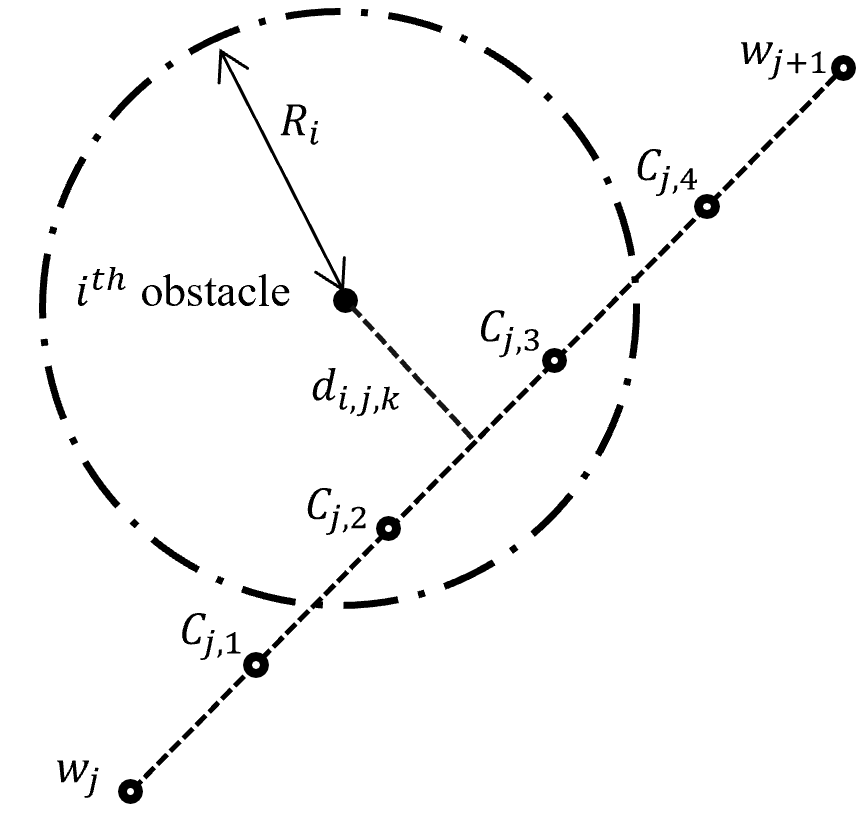}
\caption{Perpendicular distance check.}
\label{fig3}
\end{figure}

\begin{figure}[!htbp]
\centering
\includegraphics[scale=0.7]{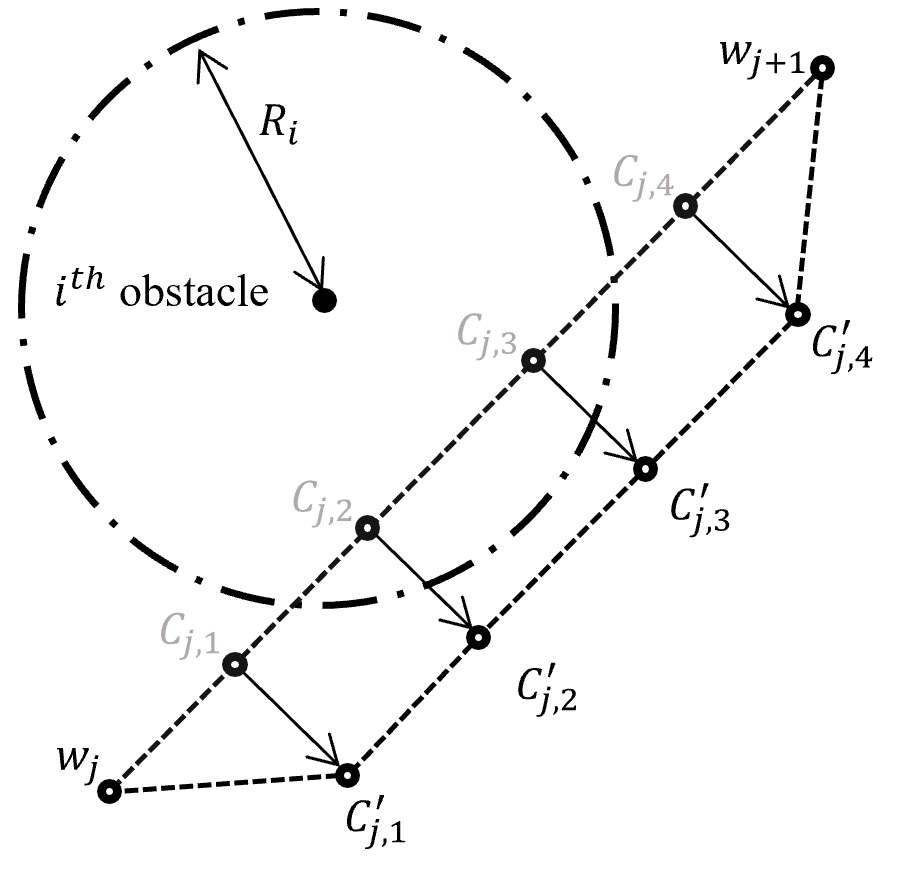}
\caption{Applying the perpendicular distance check to all the points. The points with superscript ($'$) represent the new position of the points in grey.}
\label{fig4}
\end{figure}

A resolution is to consider the perpendicular distance for each line segment from the obstacle, i.e., $d_{i,j,k}$, for $k=0,\ldots,N$, where $N$ represents the number of collocation points, $C_{j,0}=w_j$ and $C_{j,N+1}=w_{j+1}$, as illustrated in Figure \ref{fig3}, where, $d_{i,j,k}$, for $k=0,\ldots,4$, are all same for all the segments, as they are on the same straight line. Considering only perpendicular distance as the constraint $d_{i,j,k}>R_i$ leads to a very conservative approach, as it moves all points $C_{j,1}$ and $C_{j,4}$ to satisfy the constraint, as shown in Figure \ref{fig4}. This takes a lot of workspace and might lead to an infeasible optimization problem after applying the other constraints. Therefore, we only need the constraint to be applied to $C_{j,2}$ and $C_{j,3}$ in Figure \ref{fig4}, as they lie within the safety sphere. The solution we propose is to first check if the projection of $i^{th}$ obstacle position on the line segment $\overline{C_{j,k}C_{j,k+1}}$ lies on the segment or its extension. To do so, we compute
\begin{equation}\label{eq4}
    t_{i,j,k}=\frac{\left(\underline{x}_{obs,i}-C_{j,k}\right).\left(C_{j,k+1}-C_{j,k}\right)}{\left(C_{j,k+1}-C_{j,k}\right).\left(C_{j,k+1}-C_{j,k}\right)},
\end{equation}
where $\cdot$ represents the inner product. If  $0<t_{i,j,k}<1$, then the projection point is on the line segment $\overline{C_{j,k}C_{j,k+1}}$. Otherwise, it is on its extension. Accordingly, the collision avoidance constraint, illustrated in Figure \ref{fig5}, is as follows.
\begin{algorithmic}
\If{$0<t_{i,j,k}<1$ $\&$ $d_{i,j,k}<R_i$} 
    \State  Move the $C_{j,k}$ , for $k=1,\ldots,N$, such that $\overline{C_{j,l}C_{j,l+1}}$, for $l=0,\ldots,N$ is outside of the safety sphere.
\EndIf
\end{algorithmic}

\begin{figure}[!htbp]
\centering
\includegraphics[scale=0.7]{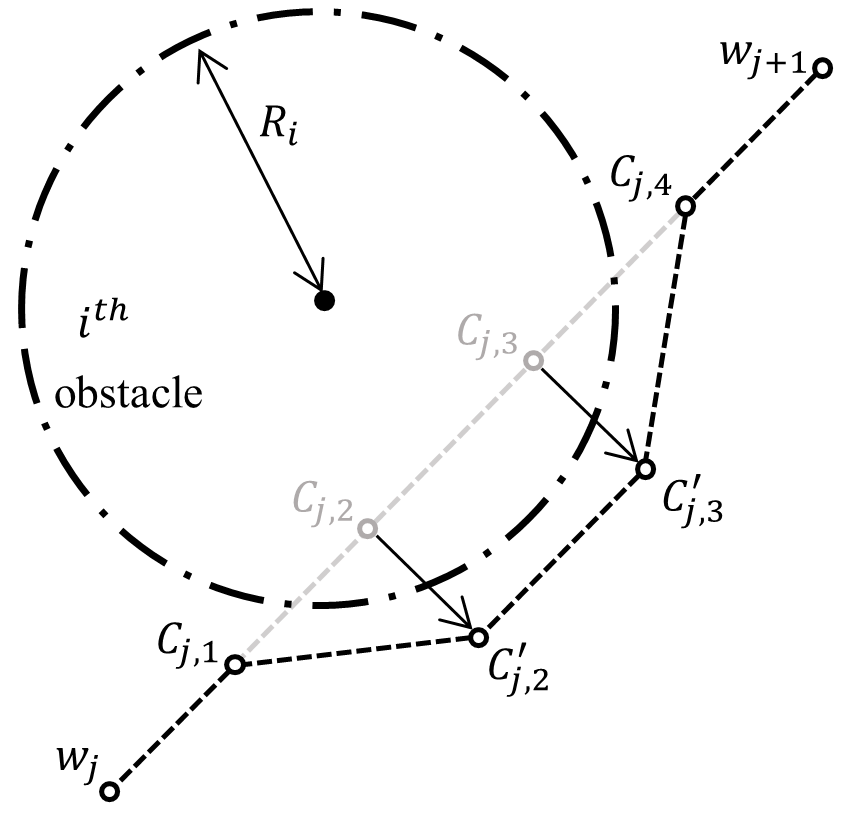}
\caption{Applying the projection point check.}
\label{fig5}
\end{figure}
This approach theoretically solves the mentioned problem, however, it imposes a practical implementation issue. This stems from the presence of the \emph{if} condition in the construction of constraints which is a common approach for obstacle avoidance constraints \cite{zhou2021ego}. Using \emph{if} condition changes the structure and size of the constraints at every iteration. This is due to the fact that by moving collocation points, the constraint might be satisfied and it is removed from the constraints matrix. More importantly, \emph{if} condition is not a smooth condition which might make a problem with the convergence of the solver. To resolve this problem, we propose a smooth constraint avoiding the use of \emph{if} condition, as
\begin{equation} \label{eq5}
    f_{i,j,k} := \frac{1}{2}\Gamma_{i,j,k}\left(R_i-d_{i,j,k}\right) \leq 0,
\end{equation}
where $\Gamma_{i,j,k}=\tanh{\left(\frac{t_{i,j,k}}{\delta}\right)-}\tanh{\left(\frac{t_{i,j,k}-1}{\delta}\right)}$ and $\delta>0$ is a small scalar. It is readily shown that for $0<t_{i,j,k}<1$, $\Gamma_{i,j,k}\approx1$. Therefore, $f_{i,j,k}=R_i-d_{i,j}$. Then the constraint $f_{i,j,k}\le0$ is equivalent to $R_i-d_{i,j,k}\le0$. On the other hand, for $t_{i,j,k}\le0$ or $1\le t_{i,j,k}$, the term $\Gamma_{i,j,k}=0$ and, hence, $f_{i,j,k}=0$. Moreover, the condition $0\leq 0$ is already satisfied in numerical solvers and it does not apply the distance constraint on the points $C_{j,k}$ and $C_{j,k+1}$. So, the constraint \eqref{eq5} encapsulates both conditions for the line segment on which the projection point lies, without using \emph{if} condition. More importantly, this constraint is smooth at points $t_{i,j,k}=0$ and $t_{i,j,k}=1$. $f_{i,j,k}$ is illustrated in Figure for $\delta=0.05$ and $R_i=1.5$.
\begin{figure}[!htbp]
\centering
\includegraphics[scale=0.3]{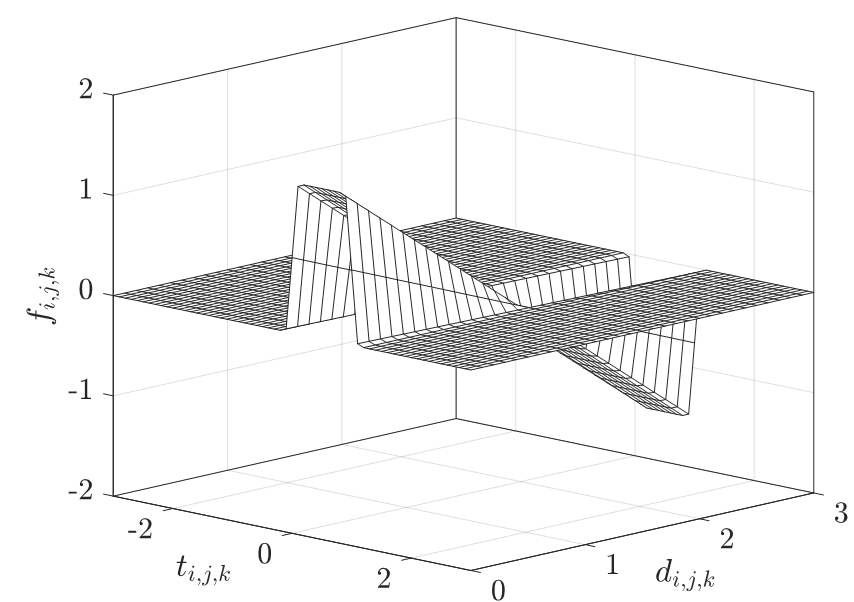}
\caption{Smooth obstacle avoidance constraint.}
\label{fig5}
\end{figure}

Another improvement to guarantee real-time feasibility with fast convergence is to make sure, the initial guess does not go through the safety sphere. To address this, we first radially move the collocation points of the initial guess out of the sphere. For each point, we check if it is inside of the sphere, if so, that point is moved in the direction of the carrying radius, i.e., connecting the centre to the point. However, this does not necessarily guarantee that the connecting line does not pass through the sphere. Therefore, we can consider the perpendicular distance of the centre to each line. However, just moving all the points according to the perpendicular distance might make lead to a conservative initial guess which is away from the optimal solution. So, we only move the lines, on which the projection point lies, not on their extension, in the perpendicular direction. This improvement might make the initial guess non-smooth and longer than the direct connecting line. So, finally, we shorten the guess, i.e., we check if the direct line between the collocation points pas through the sphere. if not, the direct line is replaced.

\subsection{Artificial Potential Field}
\begin{figure}[!htpb]
\centering
\includegraphics[width=88mm]{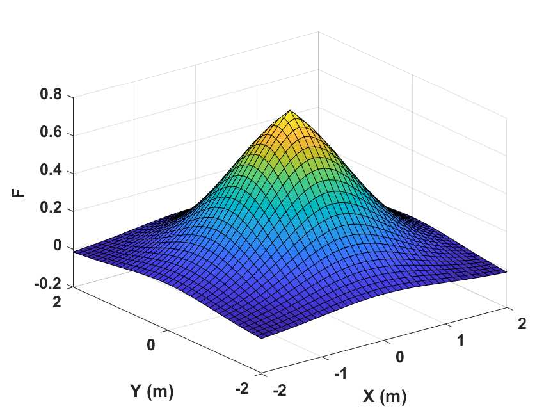}
\caption{Illustration of APF for $\alpha=1.875$ and $\eta=0.029$.}
\label{fig:apf}
\end{figure}

An APF method using sigmoid function~\cite{ren2007potential} is used to make minor corrections to the optimal control inputs to improve safety and guarantee collision avoidance in case of any unexpected delays in the computation of the optimization solution. In the case of any unexpected delay in communication or convergence issue, this ensures obstacle avoidance and safety. APF method, as illustrated in Figure~\ref{fig:apf}, creates a virtual field around the obstacle that increases or decreases in strength as the distance to it decreases or increases, respectively. In this paper, the sigmoid function is used to design the APF, as
\begin{align} \label{APF}
    F_k(t) & = \frac{1}{2}\left(1+tanh(\alpha R_k-\norm{r_k(t)})\right)-\eta
\end{align}
for $k \in \{1,...,n_o\}$, where $F_k(t)$ is the repelling force, $\alpha$ and $\eta$ are tuning parameters, and $r_k(t)={\underline{x}(t)-\underline{x}_{obs,k}(t)}$.
By solving the optimization problem \eqref{eqn:cost_function} with cost function \eqref{eqn:J_psm} with boundary conditions \eqref{eqn:boundary_conditions}, bound constraints \eqref{eqn:bound_constraints}, and obstacle avoidance constraints \eqref{eq5}, the optimal control inputs, i.e. $\underline{u}_{opt}(t)$, is obtained. Then $\underline{u}_{cmd}(t)$ is obtained by modifying $\underline{u}_{opt}(t)$ with the repelling force from APF, as
\begin{align} \label{both_methods}
    \underline{u}_{cmd}(t) & = \underline{u}_{opt}(t) + \sum_{k=1}^{n_o}F_k(t)\frac{r_k(t)}{\norm{r_k(t)}}.
\end{align}

\subsection{Parallelization}

APF \eqref{APF} can be implemented at a very high frequency due to its simplicity. However, solving the optimization problem \eqref{eqn:cost_function} is numerically expensive. It should be noted that it is not always required to solve the optimization problem at a high rate, and the trajectory can be slightly modified by APF. Therefore, we propose to implement these two components to run simultaneously in parallel, with different frequencies. In fact, the higher frequency of APF always guarntees the obstacle avoidance and, hence, the safety of the solution. The proposed parallelization is shown in a pseudocode in Table~\ref{table:pseudocode}. 

At first, the optimization problem is solved once by defining the initial guess and boundary conditions based on the initial and final UAV positions and obstacle location. Then, parallel implementation of the two methods begins in a while loop. If no optimization code is running, the control inputs at initial time $(t = 0)$ are obtained from the solution and are treated as inputs to be commanded. The optimization solution along with the co-state multipliers are used to make a new initial guess. The initial boundary condition is updated based on the new UAV position and the optimization solver begins running in the background until a new solution is computed. The APF method is used to determine the necessary repulsive corrections to the initial control inputs, obtained from the optimization solution, and it is commanded to the UAV. The process is repeated until the UAV reaches its desired position.

\begin{table}[!ht]
\caption{Pesudocode for Optimization and APF Parallelization.}
\label{table:pseudocode}
\begin{center}
\begin{tabular}{ll}
\hline
\hline
  & Initialization\\
 1. & $\underline{x}(0) \leftarrow \underline{x}_0 $ Initial condition\\
 2. & $\underline{x}(t_f) \leftarrow \underline{x}_f $ Final condition\\
 3. & Solve an initial NLP\\
 4. & $d(t) \leftarrow \norm{\underline{x}_f - \underline{x}(t)}$\\
  & Start parallelization\\
 5. & {\bf while} $d(t) > Threshold$, {\bf do}\\
 6. & \hspace*{3mm} {\bf if} NLP solved\\
 7. & \hspace*{6mm} $\underline{u}_{opt} \leftarrow \underline{u}(0)$\\
 8. & \hspace*{6mm} $\underline{x}(0) \leftarrow \underline{x}_{uav}$\\
 9. & \hspace*{6mm} $\underline{x}(t_f) \leftarrow \underline{x}_f$\\
 10. & \hspace*{6mm} Solve NLP in background\\ 
 11. & \hspace*{3mm} {\bf end} \\
 12. & \hspace*{3mm} Compute \eqref{both_methods} \\
 13. & \hspace*{3mm} Command $\underline{u}_{cmd}(t)$ to UAV dynamics \eqref{eqn:uav_dynamics}\\
 14. & \hspace*{3mm} $d(t) \leftarrow \norm{\underline{x}_f - \underline{x}(t)}$\\
 15. & {\bf end}\\
\hline
\hline
\end{tabular}
\end{center}
\end{table}

\section{Experimental Results and Discussion}
\label{sec:results}

\subsection{Experimental Setup}

\begin{figure}[!htpb]
\centering
\includegraphics[width=85mm]{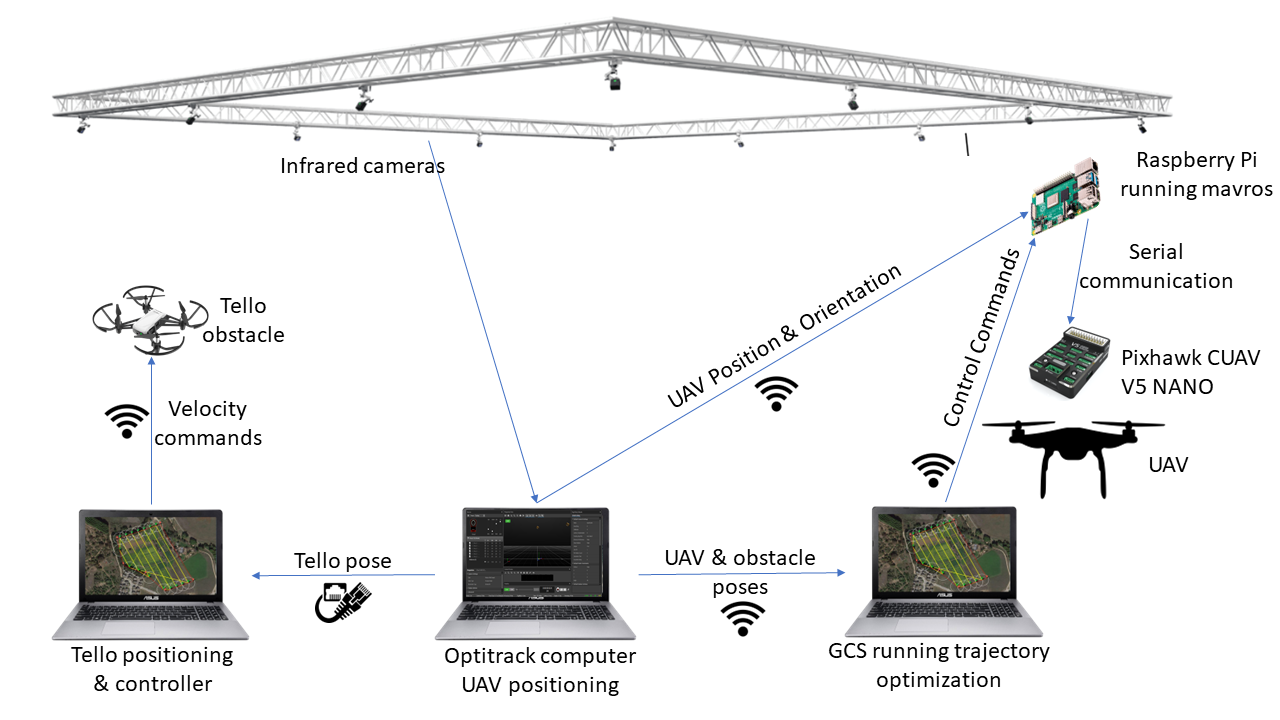}
\caption{Illustration of the UAV positioning and control network architecture in the lab.}
\label{uav_network}
\end{figure}

The setup of the UAV used for the experiments in this paper, onboard components, indoor positioning system, and communication network is illustrated in Figure~\ref{uav_network}. The UAV is a quadrotor vehicle, equipped with Pixhwak CUAV V5 Nano autopilot running the PX4 operating system. It also has a small onboard computer, Raspberry Pi Model B, running an Ubuntu server. The obstacle is a small DJI Tello drone. Positioning of the UAV and obstacle is carried out by the Optitrack system, which comprises 12 infrared cameras fixed to the ceiling at a height of around 5 $m$ and a high-performance desktop computer running Motive software. The Optitrack system is calibrated to detect the circular reflective markers fixed on the UAV and obstacle asymmetrically and determines their position and orientation with respect to a predefined FLU inertial coordinate system at a frequency of 120 Hz. The UAV's onboard computer runs roscore and acts as an interface for communication between the Optitrack system, autopilot, and the ground control station (GCS). It receives the position and orientation from the Motive software over WiFi and relays it to mavros node. The mavros node further sends it to the autopilot, which internally fuses it with the IMU's high-frequency acceleration, and angular-rate measurements and receives the PX4-estimated position, and orientation in the inertial frame to make it available to GCS. The GCS initiates ROS in Matlab and has access to all the ROS topics, including obstacle's pose, UAV's pose, and IMU data published by mavros. The trajectory optimization code runs in Matlab to determine the control inputs. The computed control inputs are commanded to the UAV by publishing them to relevant topics in mavros, which relays them to the autopilot. The Tello drone is controlled by a separate computer interfaced with both Tello and Optitrack system simultaneously by Tello's WiFi and ethernet connections, respectively. A custom ROS node is developed to interface with the Tello driver and gain access to its control.


\begin{figure*}[!htpb]
\centering
a)
\includegraphics[width=80mm]{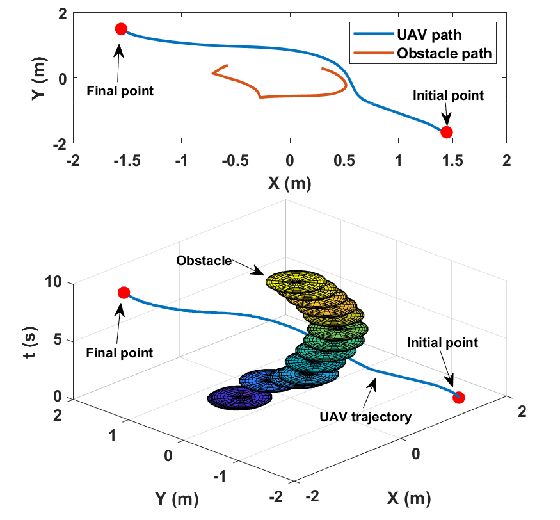}
b)
\includegraphics[width=80mm]{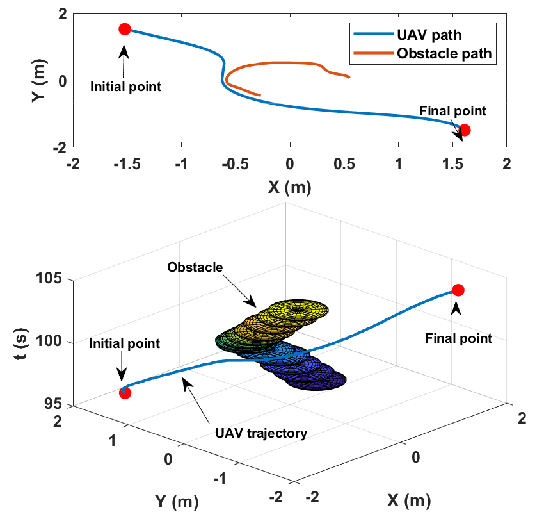}
\caption{Evolution of UAV trajectories with respect to moving obstacle: a) $t \in [0.3, 9.33]$ and b) $t \in [96.03, 103.86]$.}
\label{fig:traj_2d_3d}
\end{figure*}

\subsection{Experimental Results}
A single Tello drone is used as a dynamic obstacle and controlled to fly in a circular trajectory at an angular speed of 1.0 $rad/s$. The radius, height, and yaw angle were maintained to be constant at 0.5 $m$, 0.75 $m$, and zero $deg$, respectively. Two points $(-1.5, -1.5,0.75)$ and $(1.5,1.5,0.75)$ in the FLU frame of reference are chosen as the boundary points. The values of $\alpha$ and $\eta$ in \eqref{APF} are chosen as 1.875 and 0.029, respectively.

The UAV is commanded to take off and reach the first boundary point. Once it reached, the trajectory optimization and APF method run in parallel iteratively. In the first iteration, an optimal trajectory is computed with boundary conditions and constraints defined appropriately. To prevent abrupt accelerations to UAV, the initial velocity is constrained not to have large values. The initial control input obtained from the optimization solution was corrected using APF method and commanded to the UAV. In the second iteration, a new optimization problem is created with the new UAV and obstacle positions. By using the warmstart feature of IPOPT, the optimization solution and corresponding co-state multipliers of constraints from the first iteration are specified as the initial guess. The optimization problem runs in the background using the parfeval function in Matlab. The initial control input from the first iteration is still used to make corrections using APF method and commanded. In the subsequent iterations, it was checked if the optimization code running in the background is solved or not. If so, the routine in the second iteration is repeated. Otherwise, only the APF method is implemented. Process is repeated until the UAV reached its final point.

To ensure that the proposed approach is robust and works for different obstacle positions and approaching scenarios, the UAV is continued to fly even after reaching the final point. However, the boundary points were switched. Based on the experimental analysis, the optimization solver is so robust in computing the solution. Every time the boundary points are switched, the solution from the previous iteration is an incorrect initial guess. However, the solver is able to compute a new solution without any noticeable delays.

The experiment was carried out for 150 seconds. Throughout the experiment, the UAV was able to fly back and forth between the two boundary points without having any collision with the moving obstacle. UAV trajectories and optimal trajectories computed from 135 $s$ to the end were recorded in a video\footnote{\url{https://youtu.be/swSspfvYjJs}}. It should be noted that $y$ and $x$ axes in the figure correspond to $F$ and $-L$ axes in the FLU frame of reference. This was done to have consistency between the UAV motion in the video and the plotting axes. During the experiment, the optimization solver was able to run at a frequency of around 10 Hz. The APF method was run at around 60 Hz. Despite the differences in the frequencies of their runtime, it appears that the trajectories are getting updated in real-time without any noticeable delays. The APF method can run at a much higher frequency, but most of the time was consumed reading messages from ROS topics and publishing the control commands.

The paths and trajectories of the UAV and the obstacle from 0.3 to 9.33 $s$ and 96.03 to 103.86 $s$ are presented in Figure~\ref{fig:traj_2d_3d}. 
It can be seen that the UAV and obstacle trajectories (shown as circular disks) never intersected with each other. At the beginning of the maneuver in both cases, the obstacle was not on the line segment joining the two boundary points. Therefore, the computed optimal trajectory was a straight line. During the course of the maneuver, the obstacle started approaching the UAV and intersecting its path. The optimization solver was able to quickly compute new trajectories avoiding the obstacle. As soon as the obstacle started moving away from the UAV, the computed trajectories became almost straight again. The UAV continued reaching its final point along this straight line. Occasionally, slightly curved trajectories are computed due to the effects of the sub-optimal previous solution chosen as the initial guess. Overall, the proposed approach and the computational framework developed were shown to work effectively for real-time optimal UAV trajectory planning and dynamic collision avoidance.

\section{Conclusions} \label{sec:conclusions}
In this paper, we presented an approach for real-time optimal UAV trajectory planning in the presence of dynamic obstacles, combining trajectory optimization with APF method. The optimization problem minimized the flying time between the initial and final positions. We transcribed the problem into a nonlinear programming problem using Chebyshev pseudospectral method. The state and control histories are approximated by using Lagrange polynomials and Chebyshev nodes. More importantly, we presented a novel sigmoid-type collision avoidance constraint. Automatic differentiation of cost function and constraints were used for fast convergence. The APF method was for guaranteeing collision avoidance. We also presented a parallel architecture running at moderate and high frequencies. Indoor experiments were conducted which confirmed the effectiveness of the proposed approach.

\bibliographystyle{ieeetr}
\bibliography{references}
\end{document}